\pdfoutput=1
\documentclass[11pt]{article}

\usepackage[final]{coling}
\usepackage{booktabs}
\usepackage{times} 
\usepackage{latexsym}

\usepackage[T1]{fontenc}
\usepackage[utf8]{inputenc}

\usepackage{microtype}

\usepackage{inconsolata}

\usepackage{graphicx}

\title{Challenges in Adapting Multilingual LLMs to Low-Resource Languages using LoRA PEFT Tuning}

\author{
 \textbf{\textsuperscript{*}Omkar Khade\textsuperscript{1,2}},
 \textbf{Shruti Jagdale\textsuperscript{1,2}},
 \textbf{Abhishek Phaltankar \textsuperscript{1,2}},
 \textbf{Gauri Takalikar\textsuperscript{1,2}},
\\
 \textbf{Raviraj Joshi\textsuperscript{2,3}}
\\
\\
 \textsuperscript{1}Pune Institute of Computer Technology, Pune, India
 \\
 \textsuperscript{2}Indian Institute of Technology Madras, Chennai, India
\\
 \textsuperscript{3}L3Cube Labs
Pune, India
}

\begin{document}
\maketitle
\begin{abstract}
Large Language Models (LLMs) have demonstrated remarkable multilingual capabilities, yet challenges persist in adapting these models for low-resource languages. In this study, we investigate the effects of Low-Rank Adaptation (LoRA) Parameter-Efficient Fine-Tuning (PEFT) on multilingual Gemma models for Marathi, a language with limited resources. Using a translated Alpaca dataset with 52,000 instruction-response pairs, our findings reveal that while evaluation metrics often show a performance decline post-fine-tuning, manual assessments frequently suggest that the fine-tuned models outperform their original counterparts. The observations indicate improvements in target language generation capabilities but a reduction in reasoning abilities following language adaptation. These results underscore the need for improved evaluation methodologies and the creation of high-quality native datasets to accurately assess language-specific model performance in low-resource settings.

\end{abstract}

\section{Introduction}

The emergence of Large Language Models (LLMs) such as the Llama and Gemma series has revealed substantial abilities in managing various multilingual tasks \cite{b9, b10}. These models have shown competence in multiple high-resource languages, yet their effectiveness with low-resource languages is still a challenge that needs addressing \cite{b8, b20}. Typically, fine-tuning is used to enhance model performance in particular domains or languages. Nonetheless, this strategy has yielded inconsistent outcomes for low-resource languages \cite{b6, b1}.

Our research focuses on Marathi, which is considered a low-resource language due to the scarcity of naturally occurring training data \cite{b16, b17}. We leverage the capabilities of LoRA PEFT, a parameter-efficient approach enabling model adaptation, instead of using the classic vanilla Supervised Fine-Tuning (SFT) \cite{b2, b3}. We prefer PEFT over SFT as it works in low data scenarios, is computationally effective so more widely adopted, and avoids catastrophic forgetting due to usage of non-English data only \cite{b15, b19}. We execute this method with the Gemma models employing the Alpaca dataset, translated into Marathi. Automated assessments based on NLU and commonsense reasoning usually indicate a decline in the performance of fine-tuned models. However, human evaluations, which directly judge response quality, show that these models excel in specific contextual and cultural aspects \cite{b4, b12}.

Our study challenges the effectiveness of current evaluation methods, especially for low-resource languages \cite{b11}. We highlight how automated metrics may overlook important qualitative improvements, particularly when models produce responses that resonate with specific linguistic contexts \cite{b5}. Automated benchmarks, often based on logits, may be unsuitable for evaluating instruction-tuned models, further raising concerns about reliance on these metrics \cite{b7}. We recommend adopting more rigorous evaluation methods that better align with human judgment \cite{b19, b5}.

\section{Related Work}

Using LLMs for low-resource languages, especially Supervised Fine-Tuning (SFT), has been thoroughly researched before. SFT proves to be very effective in high-resource settings, but it falls short in low-resource languages, facing many difficulties due to the data scarcity. Methods that were curated to handle constraints of low-resource languages were used through multilingual models \cite{b1, b14}. This resulted in highlighting a performance decline, caused by cultural inconsistencies in datasets \cite{b8, b20}.

As opposed to this, some of the issues have been reduced by parameter-efficient techniques like LoRA PEFT, as they minimize the number of parameters during fine-tuning. This method signifies that computational efficiency is offered and the original model's robustness is retained, by adjusting only some of the parameters \cite{b2}. A broader study emphasized that using LoRA in low-resource settings comes with low computational overhead \cite{b3, b15}. Despite this, there remains a considerable gap for exploration when it comes to leveraging LoRA for low-resource languages on Multilingual LLMs \cite{b7}.

Existing frameworks for evaluation of low-resource languages contain limitations that need to be studied \cite{b11, b19}. Low-resource languages have cultural nuances and context-dependent accuracy embedded in them, and traditional evaluation metrics may not capture them \cite{b5, b16}. This necessitates using alternative evaluation metrics, one of them being human assessments, to corroborate model performance \cite{b4}. For example, as explored, Hindi-language tasks require cultural specificity, as it does for Marathi, our study finds \cite{b17, b4}. Thus we researched how fine-tuning methods like LoRA produce quality outputs, especially when they are used in culturally refined contexts \cite{b4, b6}.

\begin{table*}
  \centering
  \begin{tabular}{lccccc}
    \hline
    \textbf{MODEL} & \multicolumn{5}{c}{\textbf{F1 Scores}} \\
    \cline{2-6}
    & \textbf{indicsentiment} & \textbf{ai2\_arc-easy} & \textbf{arc challenge} & \textbf{indic copa} & \textbf{indic xnli} \\
    \hline
    gemma-2b       & 0.7772 & 0.4435 & \textbf{0.4240} & \textbf{0.6547} & \textbf{0.3582} \\
    gemma-2b-it    & 0.7444 & 0.4651 & 0.4043 & 0.2963 & 0.3066 \\
    gemma-2b (Mr)  & \textbf{0.9397} & \textbf{0.6048} & 0.3848 & 0.4219 & 0.1675 \\
    \hline
  \end{tabular}
  \caption{\label{tab:model_f1_scores}
    F1 Scores for Gemma1 models. 
  }
\end{table*}

\begin{table*}
  \centering
  \begin{tabular}{lccccc}
    \hline
    \textbf{MODEL} & \multicolumn{5}{c}{\textbf{F1 Scores}} \\
    \cline{2-6}
    & \textbf{indicsentiment} & \textbf{ai2\_arc-easy} & \textbf{arc challenge} & \textbf{indic copa} & \textbf{indic xnli} \\
    \hline
    gemma-2-2b       & 0.9206 & 0.6384 & 0.6463 & 0.6577 & 0.2191 \\
    gemma-2-2b (Mr)  & 0.8411 & 0.6135      & 0.5271      & 0.5764 & 0.2753 \\
    gemma-2-2b-it       & \textbf{0.9749} & \textbf{0.6851} & \textbf{0.7210} & \textbf{0.7210} &   \textbf{0.2814}   \\
    gemma-2-2b-it (Mr)  & 0.9589 & 0.6343 & 0.6374 & 0.5835 & 0.1667 \\
    \hline
  \end{tabular}
  \caption{\label{tab:model_f1_scores}
    F1 Scores for Gemma2 models. }

\end{table*}
\section{Experimental Setups}

\subsection{Dataset}
The Alpaca dataset, consisting of  52,000 instruction-response pairs originally in English, was utilized for our research. The Google translate API was used to convert the dataset's instruction, input, and output columns into Marathi so that it could be used to fine-tune Gemma models. Through this translation process, we were able to produce a sizable dataset for Marathi, which helped us build the models for a language with little resources.  The dataset that was created offered a systematic and uniform format for assessing the performance of the models on instruction-driven tasks in Marathi, making it easier to compare the base and fine-tuned variants of the Gemma models.

\subsection{Models and Fine-tuning}

For our experiments, we employed several versions of the Gemma model family \cite{b9} to assess the impact of LoRA PEFT tuning on Marathi, a low-resource language. Specifically, we worked with the following \textbf{base models}:

\begin{itemize}
    \item \textbf{gemma-2b}: A 2-billion parameter model with robust multilingual capabilities, serving as one of the baseline models.
    \item \textbf{gemma-2b-it}: An instruction-tuned variant of Gemma-2B, specifically designed to excel at instruction-based tasks.
    \item \textbf{gemma-2-2b}: An enhanced and more recent version with additional pretraining on multilingual corpora, aimed at improving performance in complex linguistic tasks.
    \item \textbf{gemma-2-2b-it}: An instruction-tuned variant of Gemma-2.2B, optimized further for multilingual and instruction-following tasks.
\end{itemize}

We fine-tuned these models using LoRA PEFT to efficiently adapt them to the Marathi language, producing the following \textbf{fine-tuned models}:

\begin{itemize}
    \item \textbf{gemma-2b (Mr)}: The fine-tuned version of Gemma-2b for Marathi using the Alpaca dataset.
    \item \textbf{gemma-2-2b (Mr)}: The fine-tuned version of Gemma-2-2b for Marathi.
    \item \textbf{gemma-2-2b-it (Mr)}: The fine-tuned version of Gemma-2-2b-it for Marathi, specialized for instruction-following tasks.
\end{itemize}

LoRA PEFT allowed us to tune a smaller subset of model parameters, which minimized computational costs while maintaining the core functionality of the Gemma models. This approach was particularly advantageous in adapting these large models to a low-resource language like Marathi, where we aimed to optimize model performance without requiring extensive computational resources.

\subsection {Evaluation}

% Our assessment emphasizes two complementary methods:

% \textbf{Automated Evaluation:} We utilize established benchmarks from AI4Bharat to assess the performance of the models on tasks such as\textbf{ IndicSentiment,  ARC-easy, ARC Challenge, Indic COPA, and Indic XNLI.} These benchmarks enable a quantitative evaluation of the models across a variety of language tasks, allowing us to compare the results with those of other multilingual models.

% \textbf{Manual evaluation:} As we used the automated metrics, we also performed thorough assessments manually, using a subset of \textbf{150 questions} from our curated sheet of questions. Then leveraging the models we generated responses for each model and each question, to ascertain which model demonstrates better performance. The questions encompassed fields like  \textbf{knowledge-based, quantitative analysis, culture and history, mathematics, science, problem-solving, scenario-based, geography and politics.} This manual evaluation revealed some important model capabilities that were previously overlooked by automated metrics, like cultural significance, linguistic patterns, nuances and capacity to follow instructions.

% By integrating both automated and manual evaluations, we achieved a more thorough understanding of model performance, pinpointing areas where fine-tuned models excel and where they may fall short.

Our assessment emphasizes two complementary methods:

\textbf{Automated Evaluation:} We utilize established benchmarks from AI4Bharat to assess the performance of the models on tasks such as IndicSentiment, ARC-easy, ARC Challenge, Indic COPA, and Indic XNLI \cite{b4}. These benchmarks enable a quantitative evaluation of the models across a variety of language tasks, allowing us to compare the results with those of other multilingual models 

\textbf{Manual Evaluation:} As we used the automated metrics, we also performed thorough assessments manually, using a subset of 150 questions from our curated sheet of questions. Then, leveraging the models, we generated responses for each model and each question to ascertain which model demonstrated better performance. The questions encompassed fields like knowledge-based, quantitative analysis, culture and history, mathematics, science, problem-solving, scenario-based, geography, and politics. This manual evaluation revealed some important model capabilities that were previously overlooked by automated metrics, like cultural significance, linguistic patterns, nuances, and the capacity to follow instructions 

By integrating both automated and manual evaluations, we achieved a more thorough understanding of model performance, pinpointing areas where fine-tuned models excel and where they may fall short .

\begin{figure*}[t]
  \includegraphics[width=\textwidth]{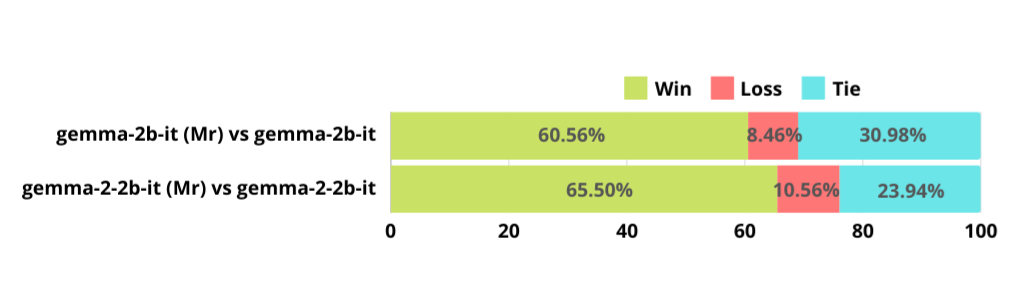}
  \caption{Manual Evaluation Performance.}
  \label{fig:experiments}
\end{figure*}

\section{Results}

\subsection{Result Discussion}

In the manual assessment of 150 questions, illustrated in Figure 1, fine-tuned versions like gemma-2-2b-it (Mr) and gemma-2b-it (Mr) showed higher win rates than their base counterparts, indicating their enhanced ability to generate contextually relevant answers in Marathi. Nonetheless, the base models occasionally generated responses in English, as depicted in Appendix Figure 2, revealing ongoing issues with language consistency that the fine-tuned models somewhat alleviated, though not completely. While the fine-tuned models performed better in most of the aspects, there were some instances where the base models matched their performance, reflecting the  intricate challenges of adapting models for low-resource languages such as Marathi.

In the evaluation of the F1 score, represented in Table 1 for gemma-1 models and Table 2 for gemma-2 models, gemma-2-2b frequently performed better than the other models in significant benchmarks, including sentiment analysis and question-answering tasks. However, fine-tuned models like gemma-2-2b-it (Mr) displayed varied outcomes, showing enhancements in certain tasks while experiencing declines in others, particularly in benchmarks like Indic XNLI and ARC Challenge. These findings highlight that even though fine-tuning can enhance performance in specific areas, it does not guarantee improvements across all tasks, underlining the necessity for more focused fine-tuning strategies for low-resource languages.

Overall, we observe a degradation in NLU and reasoning benchmarks following language adaptation. However, the adapted model performs better on the open-ended question answering dataset during manual evaluation. This suggests the need for a more comprehensive evaluation strategy and more suitable datasets to fully assess the benefits of language adaptation. While automated benchmarks indicate degradation, they may not be the ideal metric for evaluating instruction-based models. We require more effective benchmarks that can assess the reasoning capabilities of the model without relying on logit-based evaluation metrics.

\section{Limitations}

% \begin{figure*}[t]
%   \includegraphics[width=1\linewidth]{examples.PNG} \hfill
%   \caption {Responses}
% \end{figure*}

% \begin{figure}[t]
%   \includegraphics[width=1\columnwidth]{comparision chart2.png} % Increased size
%   \caption{Comparison chart.}
%   \label{fig:experiments}
% \end{figure}

While researching, we faced quite a few limitations that hindered progress. Firstly, we used a dataset that was translated, instead of fetching naturally occurring Marathi content from the web. This proved unfruitful as the translated dataset does not entirely capture the complexities of the language. Next issue we faced was of limited computational resources, which resulted in limited experimental explorations, and thwarting us from exploring a broader range of models. Another challenge pertained to comprehensively evaluating the Marathi language generation, as previous benchmarks may not understand its complexities. Furthermore, the translation process contained biases, affecting the accuracy and quality of the question-answer pairs. Lastly, high-quality Marathi evaluation datasets were scarce, limiting our abilities in judging model performance in detail, this called for more robust resources in low-resource settings.

\section{Conclusion}
To conclude, our results showcase how fine-tuning of Gemma models for Marathi using LoRA PEFT compromises performance if it is based on traditional and automated evaluation metrics. On the contrary, manual assessments indicate better performance as the fine-tuned models excel in processing culturally sound and contextually relevant responses. This necessitates the use of alternate and enhanced evaluation techniques that can successfully take into account the complex nuances of low-resource languages. 

A change needs to be made in developing more robust evaluation methods which provide more accuracy and more effective performance in low-resource settings. Moreover it is also important to perpetuate the generation of high-quality naturally occurring Marathi datasets for continued advancements in this discipline.

% This document has been adapted by Emily Allaway from the instructions for earlier ACL and NAACL proceedings, including those for NAACL 2024 by Steven Bethard, Ryan Cotterell and Rui Yan,
% ACL 2019 by Douwe Kiela and Ivan Vuli\'{c},
% NAACL 2019 by Stephanie Lukin and Alla Roskovskaya,
% ACL 2018 by Shay Cohen, Kevin Gimpel, and Wei Lu,
% NAACL 2018 by Margaret Mitchell and Stephanie Lukin,
% Bib\TeX{} suggestions for (NA)ACL 2017/2018 from Jason Eisner,
% ACL 2017 by Dan Gildea and Min-Yen Kan,
% NAACL 2017 by Margaret Mitchell,
% ACL 2012 by Maggie Li and Michael White,
% ACL 2010 by Jing-Shin Chang and Philipp Koehn,
% ACL 2008 by Johanna D. Moore, Simone Teufel, James Allan, and Sadaoki Furui,
% ACL 2005 by Hwee Tou Ng and Kemal Oflazer,
% ACL 2002 by Eugene Charniak and Dekang Lin,
% and earlier ACL and EACL formats written by several people, including
% John Chen, Henry S. Thompson and Donald Walker.
% Additional elements were taken from the formatting instructions of the \emph{International Joint Conference on Artificial Intelligence} and the \emph{Conference on Computer Vision and Pattern Recognition}.

% Bibliography entries for the entire Anthology, followed by custom entries
%\bibliography{anthology,custom}
% Custom bibliography entries only

% \begin{thebibliography}{00}

\bibliography{main}
\nocite{*}

\clearpage  % Forces a new page
\onecolumn  % Switches to single column for the appendix

\appendix
\renewcommand{\thesection}{}  % This removes the "A" from the section numbering

\section{Appendix}
\vspace{1cm}  % Add some space between section heading and figure
\textbf{Example Outputs}

\begin{figure*}[!h]  % Place figure here
    \centering
    \includegraphics[width=\textwidth, height=0.8\textheight, keepaspectratio]{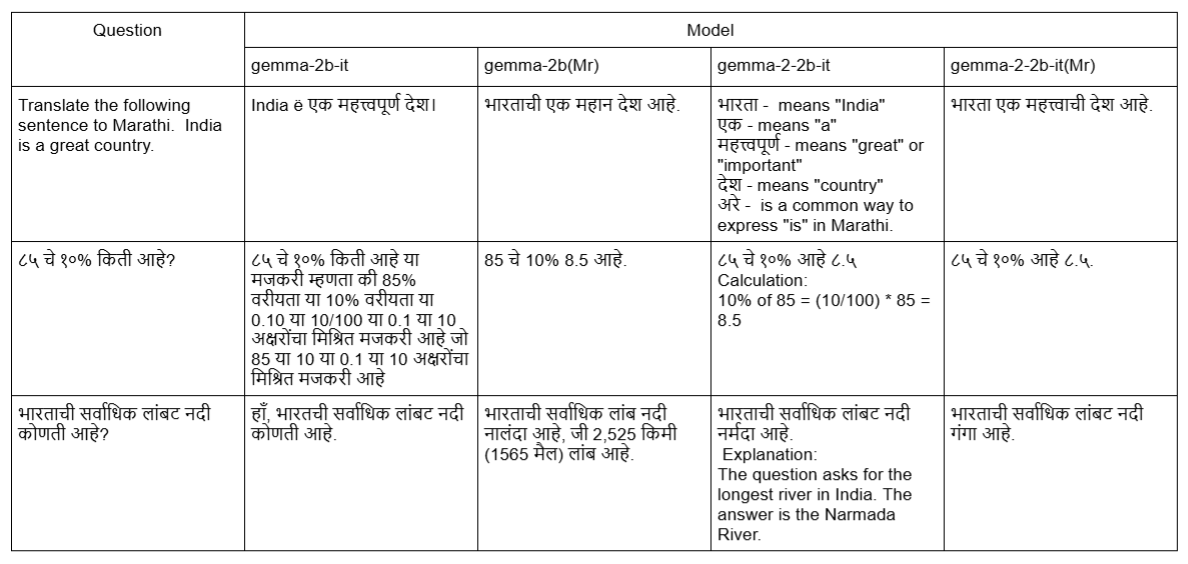}
    \caption{Responses}
    \label{fig:responses}
\end{figure*}

\twocolumn  % Returns to two-column format for subsequent pages if needed

% \begin{figure*}[t]
%   \includegraphics[width=1\linewidth]{examples.PNG} \hfill
%   \caption {Responses}
% \end{figure*}

\end{document}